%% file: main.tex
\documentclass[letterpaper, 10 pt, conference]{ieeeconf}  %

\IEEEoverridecommandlockouts                              %

\usepackage[font=footnotesize,labelfont=bf]{caption}

\usepackage{graphics} %
\usepackage{epsfig} %
\usepackage{times} %
\usepackage{amsmath} %
\usepackage{amssymb}  %
\usepackage[hyphens]{url}  %
\usepackage{graphicx} %
\usepackage{caption} %
\usepackage[style=ieee, maxbibnames=99, citestyle=numeric-comp, url = false, natbib=true]{biblatex}
\usepackage{multicol}
\usepackage{booktabs}
\graphicspath{{../}{./}{./figures/}}
\usepackage[utf8]{inputenc}
\usepackage{svg}
\usepackage{amsmath}
\usepackage[font=scriptsize,labelfont=bf]{caption}
\usepackage{graphicx}
\usepackage{xcolor}
\usepackage[justification=centering]{subfig}
\usepackage{amsmath}
\usepackage{amsfonts}
\usepackage{amssymb}
\usepackage{mathtools}
\usepackage{todonotes}
\let\labelindent\relax
\usepackage{enumitem}
\usepackage{comment}
\usepackage[per-mode=symbol]{siunitx}
\DeclareSIUnit\minute{min}
\usepackage{tikzpagenodes}
\usepackage{lineno}
\usepackage{multicol}
\usepackage[bookmarks=true]{hyperref}
\hypersetup{hidelinks,colorlinks=false}
\usepackage{mathtools}
\usepackage{setspace}
\usepackage{booktabs}
\usepackage{setspace}
\usepackage[ruled,vlined,noend]{algorithm2e}
\usepackage{optidef}
\SetKwIF{If}{ElseIf}{Else}{if}{}{else if}{else}{end if}%
\usepackage[per-mode=symbol]{siunitx}
\DeclareSIUnit\minute{min}

\DeclareMathOperator*{\argmax}{argmax}

\usepackage{mathrsfs}

\newcommand{\ph}[1]{{\textbf{#1:}}}

\title{\LARGE \bf
Risk-aware Meta-level Decision Making for \\ Exploration Under Uncertainty
}

\author{Joshua Ott, Sung-Kyun Kim, Amanda Bouman, Oriana Peltzer, Mamoru Sobue, Harrison Delecki, \\ Mykel J. Kochenderfer, Joel Burdick, Ali-akbar Agha-mohammadi
}

\begin{document}

\maketitle
\thispagestyle{empty}
\pagestyle{empty}

\begin{abstract}
Autonomous exploration of unknown environments is fundamentally a problem of decision making under uncertainty where the agent must account for uncertainty in sensor measurements, localization, action execution, as well as many other factors. For large-scale exploration applications, autonomous systems must overcome the challenges of sequentially deciding which areas of the environment are valuable to explore while safely evaluating the risks associated with obstacles and hazardous terrain. In this work, we propose a risk-aware meta-level decision making framework to balance the tradeoffs associated with local and global exploration. Meta-level decision making builds upon classical hierarchical coverage planners by switching between local and global policies with the overall objective of selecting the policy that is most likely to maximize reward in a stochastic environment. We use information about the environment history, traversability risk, and kinodynamic constraints to reason about the probability of successful policy execution to switch between local and global policies. We have validated our solution in both simulation and on a variety of large-scale real world hardware tests. Our results show that by balancing local and global exploration we are able to significantly explore large-scale environments more efficiently. 
\end{abstract}

\section{Introduction}\label{sec:intro}
\input{1_Introduction}

\section{Related Work}\label{sec:rel_works}

\input{2_RelatedWork}

\section{Problem Formulation}\label{sec:prob_form}
\input{3_ProblemFormulation}

\section{Risk-aware Meta-level Decision Making}\label{sec:riskawareMLDM}

\input{4_RiskAwareMLDM}

\section{Results}\label{sec:results}
\input{5_Results}

\section{Conclusion}\label{sec:conclusion}
\input{6_Conclusion}

\section{Acknowledgments}\label{sec:acknowledgements}
\input{7_Acknowledgements}

\renewcommand*{\bibfont}{\footnotesize}
\printbibliography

\end{document}

%% file: 1_Introduction.tex
A key challenge in deploying artificially intelligent systems in real-world environments is sequential decision-making under uncertainty, particularly when dealing with large-scale tasks that necessitate a delicate balance between exploration and exploitation. This uncertainty can manifest in numerous ways, such as localization errors, hazard identification, communication delays, motion execution, and noisy sensor measurements \cite{otsu2020supervised}. In tasks that involve thorough exploration of unknown environments, an autonomous agent must account for resource limitations, the unpredictability of the task area, and the necessity to promptly and efficiently react to emergent risks \cite{agha2018slap, agha2021nebula}. Consequently, quantifying uncertainty and integrating it into strategic planning is of vital importance.

\begin{figure}[t]
\centerline{\includegraphics[width=1\columnwidth]{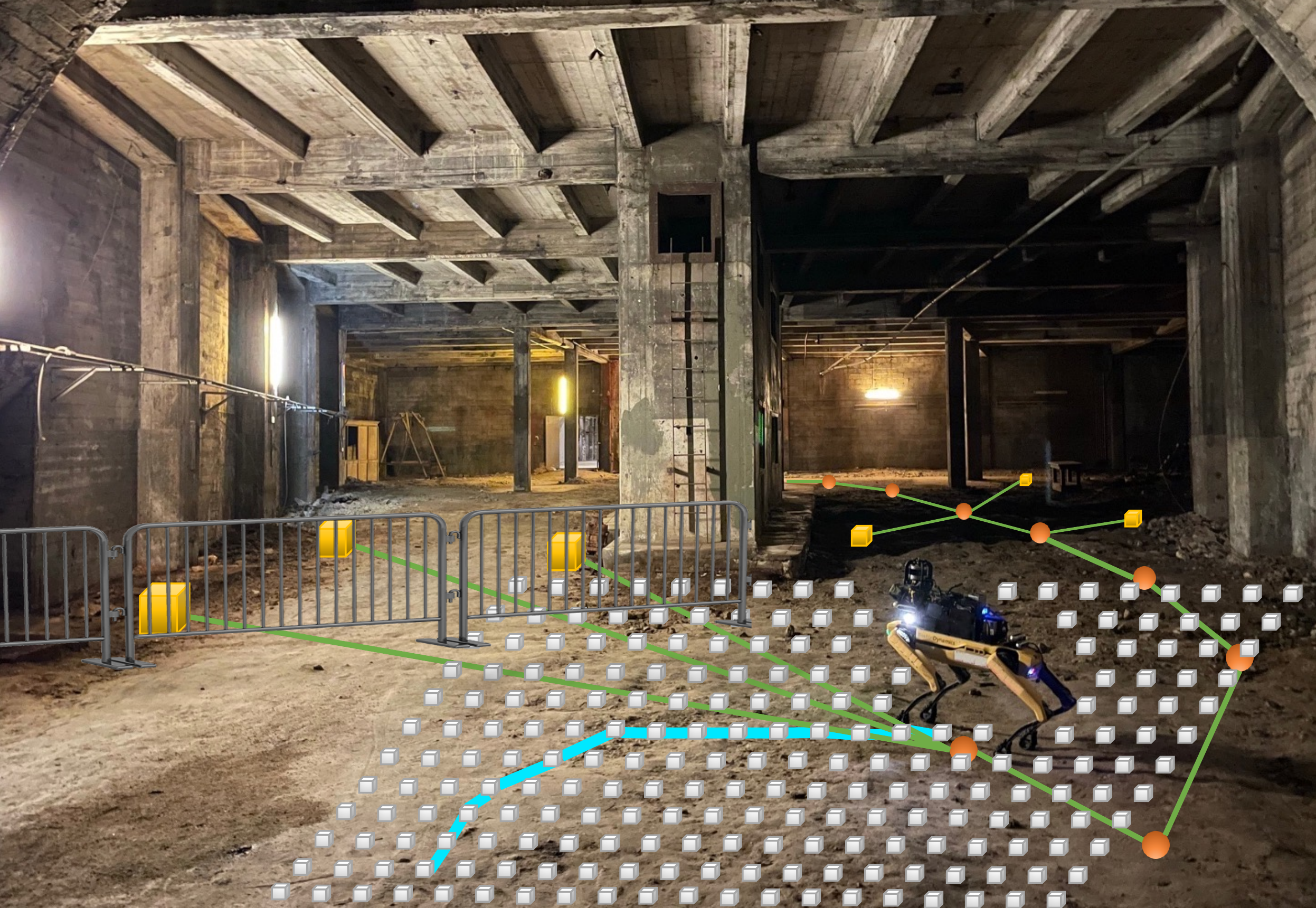}}
\caption{An example of how meta-level decision making is used to balance the tradeoffs between local and global exploration. Global planning is conducted over the global IRM as shown by the orange spheres (breadcrumbs) connected by green edges. The yellow cubes represent global frontiers. Local planning is conducted over the local IRM (grid represented by gray cubes). In this case, the robot is currently following the global planner to reach the global frontiers shown by the yellow cubes on the left. However, the global frontiers are unreachable (behind the fence), requiring the robot to choose between following the local planner (cyan line) or relocating to a different global goal like those shown in the back right of the image.} %
\label{mldm}
\end{figure}

Consider an autonomous agent, as illustrated in Figure \ref{mldm}, assigned to explore a large and uncharted environment. Constraints such as finite energy supply and noisy sensor observations affect the agent's ability to completely cover the area. Thus, prioritization and risk evaluation become essential components of the task, along with the continuous challenges that arise from real-world deployment.

There are two primary challenges in this context: the planning of coverage paths across a large scale, and the risk-aware planning that accommodates the hazards present in the environment. These challenges require an intelligent system that can accommodate these two extremes using limited computational resources. A common strategy is to divide and conquer, where a large problem is reduced to smaller, more manageable sub-problems. While this approach helps make the problem computationally tractable, it often leads to suboptimal solutions due to the nature of the problem division.

To navigate these challenges, we introduce a methodology that leverages spatial and temporal approximations to facilitate planning with an online, real-time solver. We decompose the agent's belief space into two key components: the global and local Information Roadmap (IRM). This dual representation is necessary to handle the expansive spatial and temporal scope of the task, shown in Figure \ref{mldm}. We then use Receding Horizon Planning (RHP) over this hierarchical structure in real-time, and use meta-level decision making to balance local risk-resilience and global reward-seeking objectives.

Previous work has shown the effectiveness of hierarchical planning in exploring vast unknown environments \cite{kim2021plgrim, peltzer2022fig, bouman2020autonomous, cao2021tare, dang2019graph, bircher2018receding}. However, traditional divide-and-conquer methods may result in inefficiencies due to problem bifurcation and lack of environmental knowledge. These inefficiencies are addressed in our proposed methodology. We select a coverage policy (either local or global) based on its maximum expected discounted return, weighted by the probability of successful execution.

By estimating the success probability and reward for successful plan execution, our method aims to select the most rewarding coverage policy at any given point in time. This allows for flexible exploration and ensures maximum coverage while considering traversal risk. For instance, in a scenario where a robot explores a long narrow corridor leading to a dead-end, our method provides the ability to recompute and select a more valuable policy, whether local or global, once the dead-end is discovered. This adaptability is a key feature of our approach.

\begin{figure}[t]
\centerline{\includegraphics[width=0.32\textwidth]{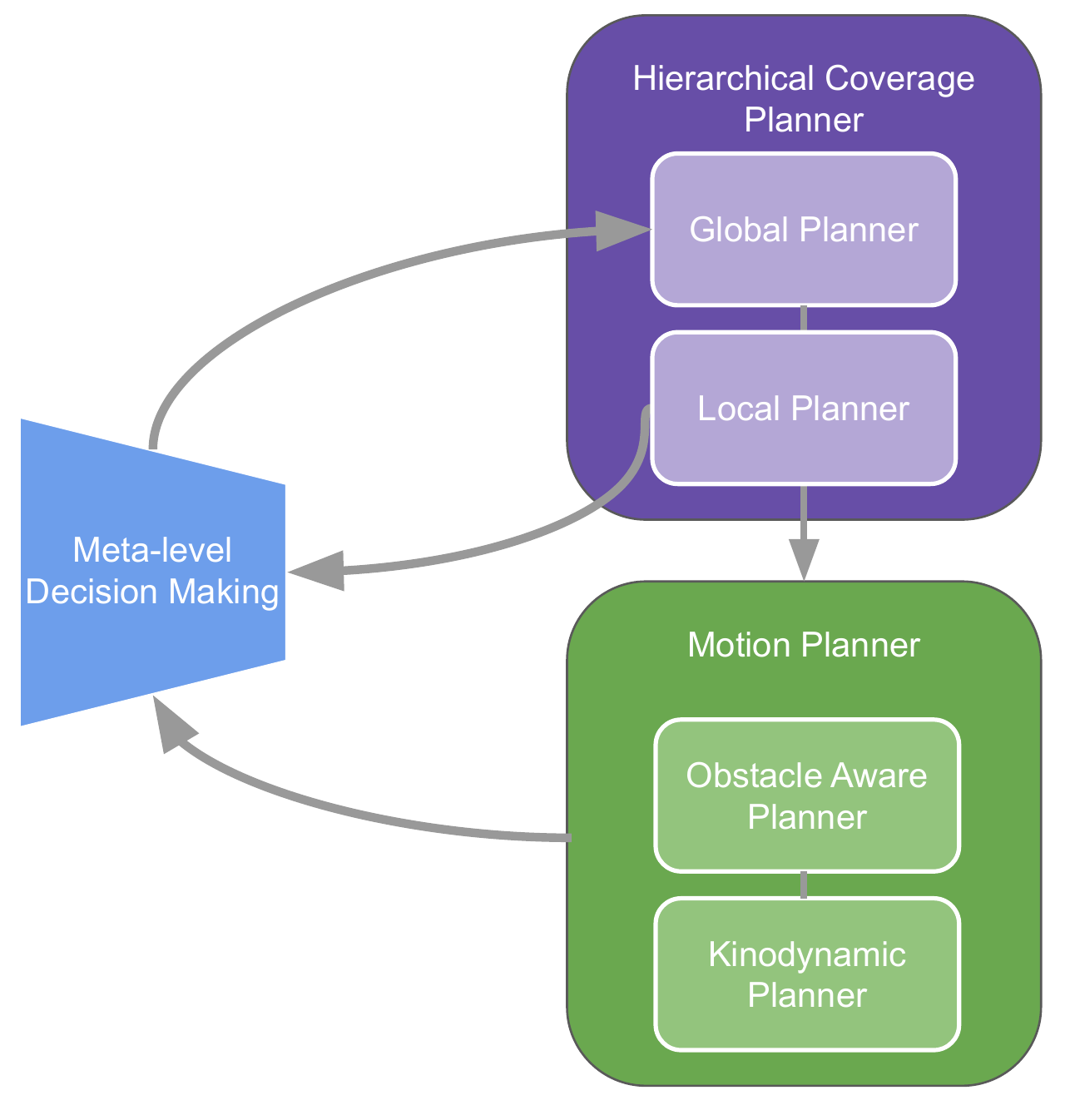}}%
\caption{Illustration of how meta-level decision making fits into the planning and motion execution pipeline as well as how information is shared amongst these modules \cite{kim2021plgrim, peltzer2022fig} \cite{fan2021step, fan2021learning}. After solving for the local or global coverage policy, the high-level goals are sent to a low-level motion planner which plans a higher resolution path using $A^*$ \cite{hart1968formal} which is then sent to the kinodynamic planner \cite{camacho2013model, fan2021step, fan2020deep}. %
}
\label{mldm_flow}
\vspace{-2mm}
\end{figure}

Our results demonstrate that in environments with complex terrain, this risk-aware meta-level decision making approach enables the autonomous agent to identify alternative paths and avoid risky regions, resulting in more efficient exploration. We rely on the hierarchical coverage planner formulation used by \citet{kim2021plgrim}, and the high-level interaction of the different planning structures is shown in Figure \ref{mldm_flow}.

The contributions of our work are threefold.

\begin{enumerate}
    \item We formulate a meta-level decision making problem for hierarchical coverage planning.
    
    \item We introduce a risk-aware meta-level decision making algorithm that reasons about the expected discounted reward and probability of successful policy execution to adaptively switch between local and global policies.
    
    \item We demonstrate our algorithm on physical multi-robot teams during large-scale exploration missions. We have deployed our solution in a variety of environments ranging from the Los Angeles Subway Station to the Kentucky Underground Limestone Mines. The proposed solution also served as the decision making system for switching between the local and global policies for team CoSTAR's entry in the Final Circuit of the DARPA Subterranean Challenge \cite{agha2021nebula}.
\end{enumerate}  

The remaining sections of this paper are structured as follows. Section 2 discusses related work. Section 3 formalizes the hierarchical coverage planning problem. Section 4 introduces the risk-aware meta-level decision-making algorithm. Section 5 presents the results from our deployment on physical robotic systems. We show that our proposed meta-level decision-making methodology, by using risk-aware information to guide policy switching, allows for more efficient and higher quality exploration in large-scale environments.

%% file: 2_RelatedWork.tex
The related literature encompasses themes of meta-level control, meta-learning, and meta-reasoning aimed at enhancing decision-making for specific goals \cite{corkill1983use, barber2000strategy}, vital for complex autonomous systems \cite{tambe1997towards}. Meta-level control optimizes performance by strategic activity selection \cite{raja2007framework}, balancing computation and solution quality, known as meta-reasoning \cite{svegliato2020model, dannenhauer2014toward, cheng2013multiagent, aiello1991reasoning}.

\ph{Hierarchical Reinforcement Learning} In hierarchical reinforcement learning, autonomous agents benefit from high-level policy switching between low-level policies for safety and efficiency \cite{cao2020reinforcement, vilalta2002perspective, schweighofer2003meta, finn2017model, vanschoren2019meta, yel2021meta, chan1993toward, finn2019online}. Others achieved better outcomes with hierarchical structures for autonomous driving behavior \cite{guo2021hierarchical}, or multiobjective approaches for vehicle engine control \cite{wray2021engine}. These methods focus on offline policy switching, contrasting with our focus on online policies for exploring unknown environments \cite{kim2021plgrim}.

\ph{Coverage Planning} Coverage planning uses approximate POMDP solvers for extended temporal and spatial scenarios. Integration of policy search under Gaussian beliefs \cite{indelman2015planning, martinez2009bayesian} and expansions using the POMCP solver \cite{lauri2016planning} have been applied. Frontier-based exploration \cite{umari2017autonomous}, risk-aware navigation \cite{kim2019bi}, and superior hierarchical POMCP frameworks \cite{vien2015hierarchical} demonstrate the effectiveness of hierarchical planners in large-scale exploration \cite{kim2021plgrim}, highlighting the benefits of a global-local approach.

Our work, set in the context of these existing studies, centers on risk-aware considerations for online planning. %
Specifically, our approach benefits from the capability to assess the advantages of switching between local and global policies using risk-aware information, an aspect not addressed by earlier works.

%% file: 3_ProblemFormulation.tex
The coverage planning problem formulation is based on that of \citet{kim2021plgrim}. Our risk-aware meta-level decision making contribution builds upon this formulation.

\subsection{\textbf{Hierarchical Coverage Planning}}
In unknown space coverage domains, we do not have strong priors about the parts of the world that have not yet been observed. Hence, knowledge about the coverage and risk state of the world %
at runtime is incomplete and often inaccurate. Thus, in such domains, Receding Horizon Planning (RHP) has been widely adopted \cite{bircher2016receding}. 

\ph{Global and Local Policies}
We decompose the policy into local and global coverage policies: $\pi^\ell$ and $\pi^g$, which are solved for over the local and global graphs $G^{\ell}$ and $G^g$ respectively. The local and global policies $\pi_{t:t+T^\ell}^\ell(b^\ell)$ and $\pi_{t:t+T^g}^g(b^g)$ are therefore given by 
 \begin{equation}
 \pi_{t:t+T^\ell}^\ell(b^\ell) = \argmax_{\pi^\ell \in \Pi_{t:t+T^\ell}^\ell} \, \mathbb{E} \sum_{t'=t}^{t+T^\ell} \gamma_{\ell}^{t'-t} r^\ell(b_{t'}^\ell, \pi^\ell(b_{t'}^\ell))
  \label{eq:llp_optimization}
\end{equation}
\begin{equation}
  \pi_{t:t+T^g}^g(b^g) = \argmax_{\pi^g \in \Pi^g_{t:t+T^g}} \, \mathbb{E} \sum_{t'=t}^{t+T^g} \gamma_g^{t'-t} r^g(b^g_{t'}, \pi^g(b^g_{t'}))
  \label{eq:glp_optimization}
\end{equation} respectively where $\gamma \in (0, 1]$ is a discount factor for the future rewards, $\Pi_{0:\infty}$ is the space of possible policies, and $r(b,a)=\int_s R(s,a)b(s)\mathrm{d}s$ denotes the expected reward of taking action $a$ at belief $b$. The \emph{discounted} utility function is defined as the expected reward of following policy $\pi^{a_t}$, starting from belief $b$: \begin{equation}
    U(b_t; \pi^{a_t}) = \mathbb{E} \sum_{t'=t}^{t+T} \gamma^{t'-t} r(b_{t'}, \pi^{a_{t}}(b_{t'})). 
    \label{eq:U}
\end{equation} \textcolor{black}{The overall policy $\pi \in \Pi$ is constructed by combining the local and global policies. The way in which we combine the local and global policies, or equivalently, how we switch between them, is the key focus of our work.}

\subsection{\textbf{Hierarchical Coverage Policy Execution}}
Once we have solved for a coverage policy, we must translate that high-level policy into action execution commands for the robot to follow. During the coverage plan execution phase, low-level motion planning uses high-fidelity risk information to safely guide the robot's actions. In this work, we consider a hierarchical motion planning framework, which has been widely adopted in many autonomous navigation applications \cite{fan2021step, pajaziti2014slam, da2020novel}.

To execute a particular policy, the local and global coverage policies are passed to lower-level motion controllers as shown in Figure \ref{mldm_flow}. First, the obstacle-aware planner takes in the high-level policy and plans a path $x_{A^*}$ using the $A^*$ algorithm \cite{hart1968formal}. This provides a higher resolution path that follows the goals set by the local or global policies. Then $x_{A^*}$ is used as a reference by the kinodynamic planner which plans a path $x_{kino}$ accounting for the kinodynamic constraints of the robot. Further information on the kinodynamic planner is provided by \citet{fan2021step, fan2020deep}.

%% file: 4_RiskAwareMLDM.tex
Our contribution is focused on switching between the local and global policies using risk-aware information. %
The objective in risk-aware unknown environment exploration is to cover, or explore, as much of the environment as possible while minimizing action costs and avoiding risky policies that could jeopardize the safety of the robot. The challenge lies in balancing between exploring a local area, which will provide some immediate coverage reward, or relocating to a different global frontier, which could provide more coverage reward in the future while also requiring more travel time.

\begin{figure}[t]
\centerline{\includegraphics[width=0.5\textwidth]{  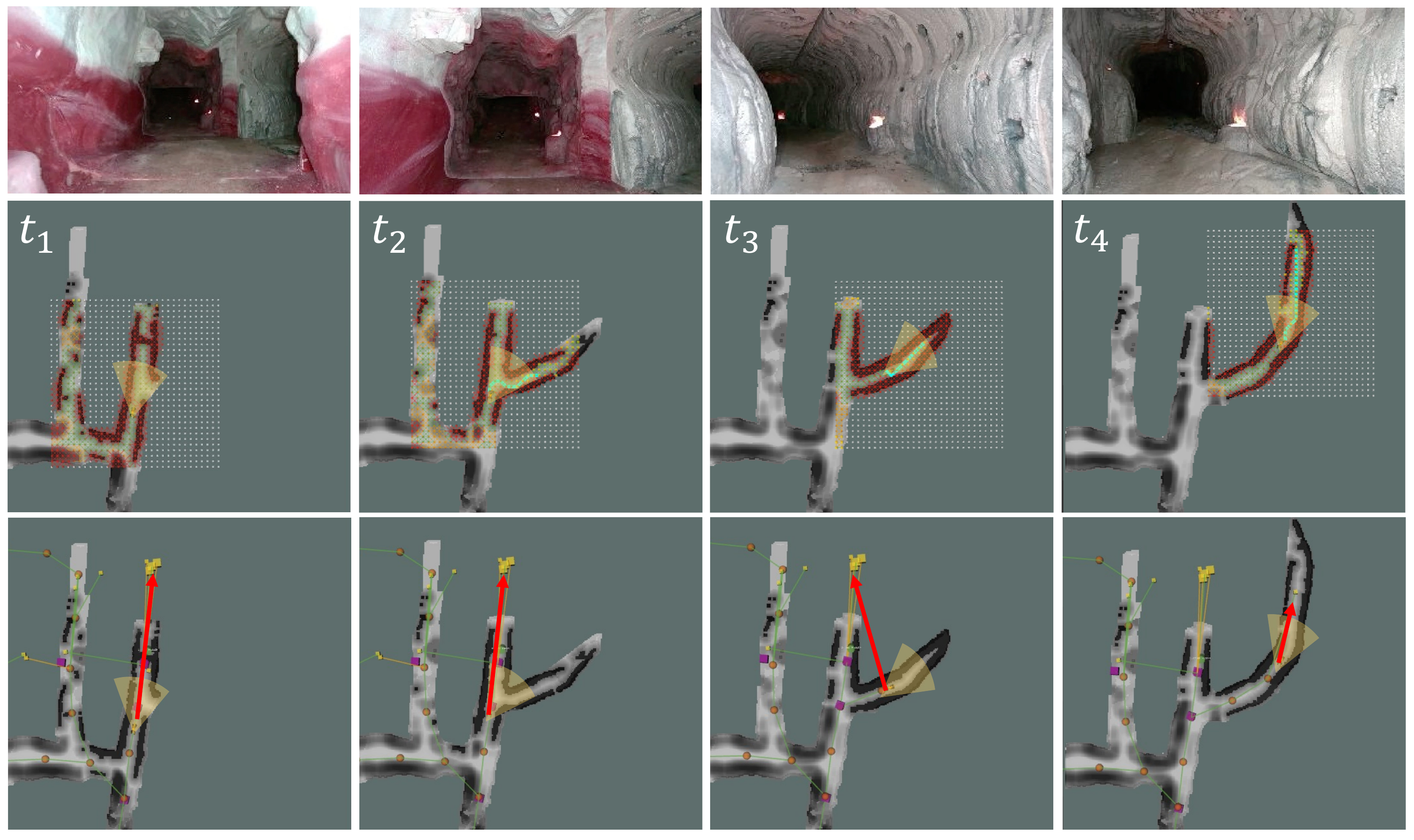}}
\caption{Example of global to local switching from a real world mission during the DARPA Subterranean Challenge. %
The first row displays the front view from cameras on the robot. The second row shows the local IRM (discretized grid), local coverage planner path (cyan line), and the direction of the robot (shaded yellow sector). The bottom row shows the global IRM (edges are green, frontiers are yellow, and previous robot locations are brown), the global goal (red arrow), and the direction of the robot (shaded yellow sector). The four columns represent four different time instances. %
}
\label{glp_override}
\vspace{-2mm}
\end{figure}

\ph{Risk-aware Information} %
By reasoning about the probability of successful policy execution, we are able to balance safety and performance in a principled manner. Incorporating this probability is one of the key aspects of our risk-aware solution. Our decision problem can therefore be stated as: 

\begin{equation}
\max_{\pi^{a_t} \in \left\{\pi_{t:t+T^g}^g, \, \pi_{t:t+T^\ell}^\ell \right\}} \, \hat{P}(\pi^{a_{t}}) U(b_t; \pi^{a_t}).
\label{eq:Q_hat}
\end{equation} 
\textcolor{black}{In other words, we are selecting the policy at time $t$ that has the greatest coverage reward if the policy is successful, multiplied by the probability that the policy is able to be successfully executed.}

\subsection{\textbf{Probability of Successful Policy Execution}}
As previously mentioned, one of the key challenges with hierarchical coverage planning is fusing the divided information between different levels to allow for more well informed decision making. We achieve this fusion of information by reasoning about the probability that a policy is able to be successfully executed. %
To do so, we leverage information across all layers of the hierarchical structure shown in Figure \ref{mldm_flow}. Specifically, we use the history of the local and global graphs, traversability risk, and kinodynamic feasbility to determine the probability of successful execution with respect to each of these components $P_{\text{G}}(\pi)$, $P_{W_r}(\pi)$, and $P_{\text{kino}}(\pi)$ respectively. An exact expression for each of these terms is problem specific and will be provided in further detail below; however, the relationship between these terms is a general idea that is widely applicable to a variety of problems. The overall probability of successful policy execution is then given by the proportionality:
\begin{align}
    \hat{P}(\pi) \propto P_{\text{G}}(\pi) P_{W_r}(\pi) P_{\text{kino}}(\pi).
    \label{eq:phat_prop}
\end{align} Because our goal is to maximize Equation \ref{eq:Q_hat}, we drop proportionality constants because they will not affect the maximum.

\begin{figure*}[t]
\centerline{\includegraphics[width=1\textwidth]{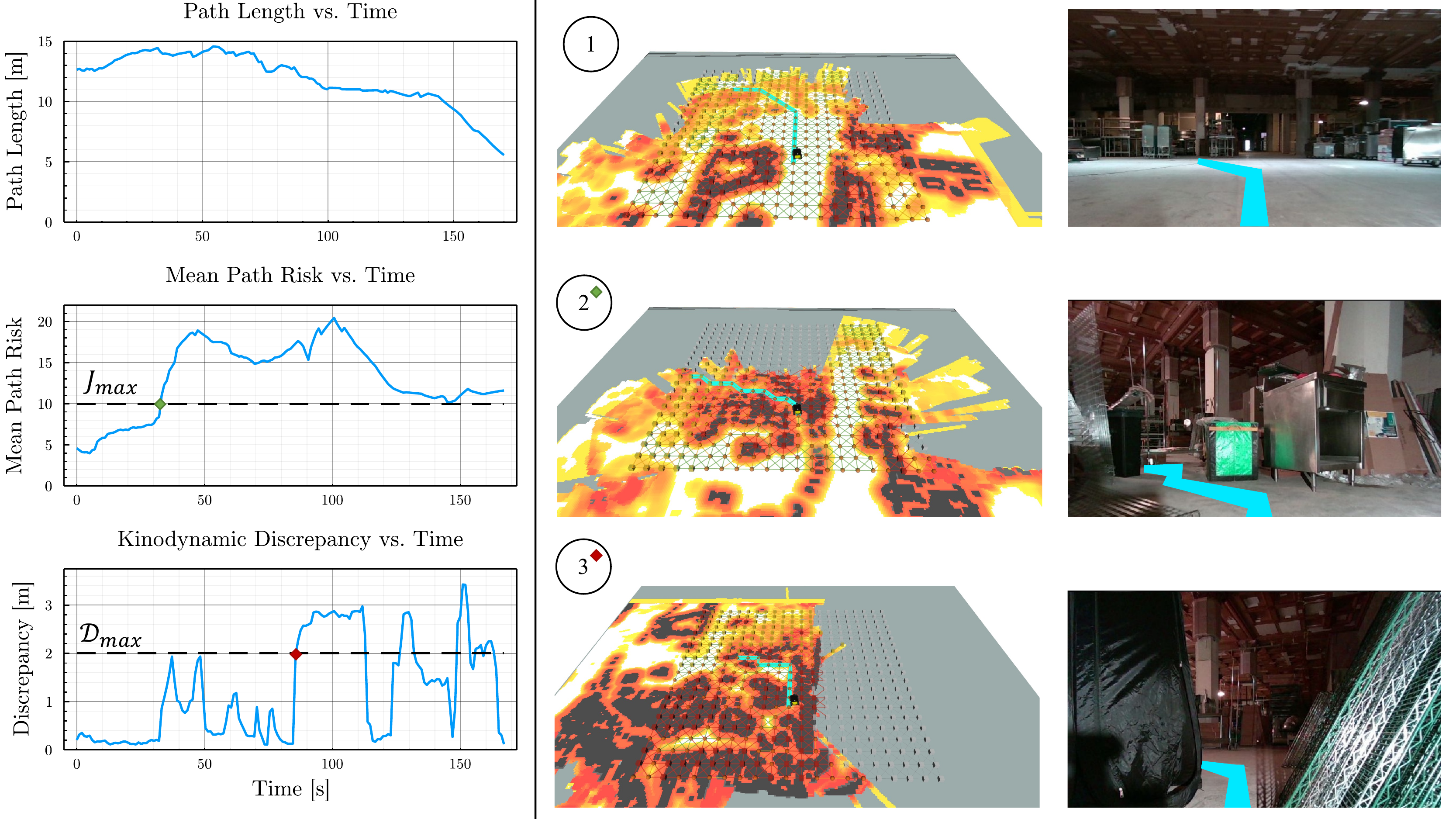}}
\caption{Example highlighting the motivation for local to global switching from a real world test conducted in the LA Subway. This test was run without meta-level decision making onboard to demonstrate the necessity of using the algorithm. Initially, the robot is in an open space as shown in the center and right images of row 1. The robot then enters a cluttered environment that has high reward (since it has not been covered yet) and very high traversability risk (row 2). Obstacles are shown in black, areas of high traversability risk are shown in red, and white indicates open space. By replaying this data with meta-level decision making running, we can see that at this point the algorithm would recommend switching to a global frontier rather than continuing with the risky local plan; however, since the algorithm was not running, the robot continues to explore the constrained environment. %
The green and red diamonds indicate the points in time where these events would have occurred if the algorithm had been running in real time on the robot. The robot's local plan is indicated by the cyan line.}
\label{mlp_override}
\vspace{-5mm}
\end{figure*}

\ph{Environment History} The probability of successful policy execution based on the history of the local and global graphs $P_{\text{G}}(\pi)$ captures the interaction between the uncertainty in the world representation and the ability to find successful policies over these representations. The local and global graphs $G^{\ell}$ and $G^g$ can vary from one planning step to the next based on new sensor information. As a result the local and global policies will change from one iteration to the next and a policy that was previously successful may no longer be feasible. Capturing this history provides a level of confidence in our respective policies over time which leads to the probability of successful execution based on the history of the environment.

\ph{Traversability Risk} 
The probability of successful policy execution based on the current traversability risk of the environment $P_{W_r}(\pi)$ is perhaps the most intuitive component of the overall $\hat{P}$ proportionality. Riskier environments will have a lower probability of successful execution since more complex terrain naturally includes greater uncertainty in the outcome of specific action executions. \textcolor{black}{Further detail on how we estimate this probability is provided in the next subsection.} 

\ph{Kinodynamic Constraints} The probability of successful policy execution based on the robot's kinodynamic constraints $P_{kino}(\pi)$ captures the feasibility of execution based on the dynamics of the robot. For example, a policy may set a goal that is incompatible with the dynamic constraints of the robot (i.e. the robot is not physically capable of moving along the requested path).  %

\subsection{\textbf{Practical Execution Probability Implementation}}
While the general idea of Equation \ref{eq:phat_prop} is applicable to a variety of robotic applications, we provide a more concrete definition here for the specific problem of risk-aware exploration. While there is no closed form analytical expression for this probability, we can still make use of the fact that the probability that a path can be successfully executed is correlated with many aspects of the environment. \textcolor{black}{For example, a possible metric to quantify the environment history is the consistency of policy generation (i.e. the number of previous policies found within a given time window), which we denote $h$.} In hazardous environments, it is more challenging to find potential policies and therefore $h$ will be lower. That is, $P_G(\pi) \propto h$. Similarly, the traversability risk accumulated by following a particular policy is given by \begin{equation}J(\pi^{a_t}) = \sum_{i=1}^{T^{a_t}-1} \rho_{i,i+1} \end{equation} where $\rho_{ij}$ represents the traversability risk associated with traversing from node $n_i$ to node $n_j$ along edge $e_{ij}$ when following either the local or global policy chosen by action $a_t$ (recall that a local or global policy is made up of actions which correspond to edge traversals in the IRM graph structure and a particular policy will have $T^{a_t}$ nodes corresponding to $T^{a_t}-1$ edges to visit based on the respective horizon). The probability of successful policy execution based on the current traversability risk of the environment can therefore be represented as $P_{W_r}(\pi) \propto 1/J(\pi)$.

\begin{algorithm}[t]
\setstretch{1.15}
\SetAlgoLined
\DontPrintSemicolon
\KwResult{An action that selects a coverage policy} 
    \For{$a_t \in \{a^{\ell} , a^{g} \}$}{
    \text{solve for } $\pi^{\ell}$, $\pi^g$, $U(b_t; \pi^{a_t})$ [Eq. \ref{eq:llp_optimization}, \ref{eq:glp_optimization}, \ref{eq:U}]
    $\hat{P}(\pi^{a_t}) \leftarrow h / (J(\pi^{a_t}) \mathcal{D}(x_{kino}, x_{A^*}))$ [Eq \ref{eq:P_hat}]
    }
    $a_t \leftarrow \argmax_{a_t} \hat{P}(\pi^{a_t}) U(b_t; \pi^{a_t})$\\
    \uIf{$J(\pi^{a_t}) > J_{max} \lor \mathcal{D}(x_{kino}, x_{A^*}) > \mathcal{D}_{max}$}{
        \uIf{$a_t = a^{\ell}$}{
            {\Return $a^{g}$}
            }
        \Else{\Return $a^{\ell}$}
        }
  \Else{\Return $a_t$}
\caption{Risk-aware meta-level decision making}
\end{algorithm}

\ph{Traversability Model} We use a traversability risk assessment model to determine $\rho_{ij}$ which is based on the tail risk assessment using the Conditional Value-at-Risk (CVaR) that previous work has shown to perform well \cite{fan2021step}. It is important to note here that our sequential decision making relies on the traversability risk assessment of the environment which has been the focus of extensive previous work \cite{fan2021step, fan2021learning}. We leverage this work to achieve risk-aware decision making; but traversability risk assessment is not the focus of this contribution. 

\ph{Kinodynamic Discrepancy} The discrepancy between the $A^*$ path $x_{A^*}$ and the kinodynamic path $x_{kino}$ provides a metric of kinodynamic feasibility. A path that has lower discrepancy is more kinodynamically feasible. Similarly, a large discrepancy indicates that the high-level policy is setting a goal for the robot that may require aggressive motion. This discrepancy could also indicate that there is some undetected anomaly such as a rock underneath the robot or some other general system failure. We define $\mathcal{D}$ as the discrepancy between the kinodynamic path $x_{kino}$ and the $A^*$ path $x_{A^*}$: \begin{equation}\mathcal{D}(x_{kino}, x_{A^*}) = \sum_{i=1}^{N} |x_{i,A^*} - x_{i,kino}|,\end{equation} where $N$ is the length in nodes of each path. We then have that $P_{kino}(\pi) \propto \frac{1}{\mathcal{D}(x_{kino}, x_{A^*})}$. Combining these factors gives: \begin{equation}\hat{P}(\pi) \propto h \cdot \frac{1}{J(\pi)} \cdot \frac{1}{\mathcal{D}(x_{kino}, x_{A^*})}. \label{eq:P_hat}\end{equation}

\ph{Algorithm} The overall implementation is provided in Algorithm 1. We can see that as the traversal risk $J$ increases, the probability of successful execution decreases. Similarly, if the kinodynamic path differs significantly from the requested path ($\mathcal{D}$ increases), then the probability of successful execution will decrease. These factors allow us to achieve real time risk-aware meta-level decision making.

%% file: 5_Results.tex
There are two main scenarios where meta-level decision making stands out in hierarchical coverage planning. \textcolor{black}{The first situation occurs when the local coverage policy is more valuable than the global coverage policy, prompting a switch from global to local planning, and the second situation occurs when switching from local to global planning is more valuable.} %

\begin{figure}[t]
\centerline{\includegraphics[width=0.5\textwidth]{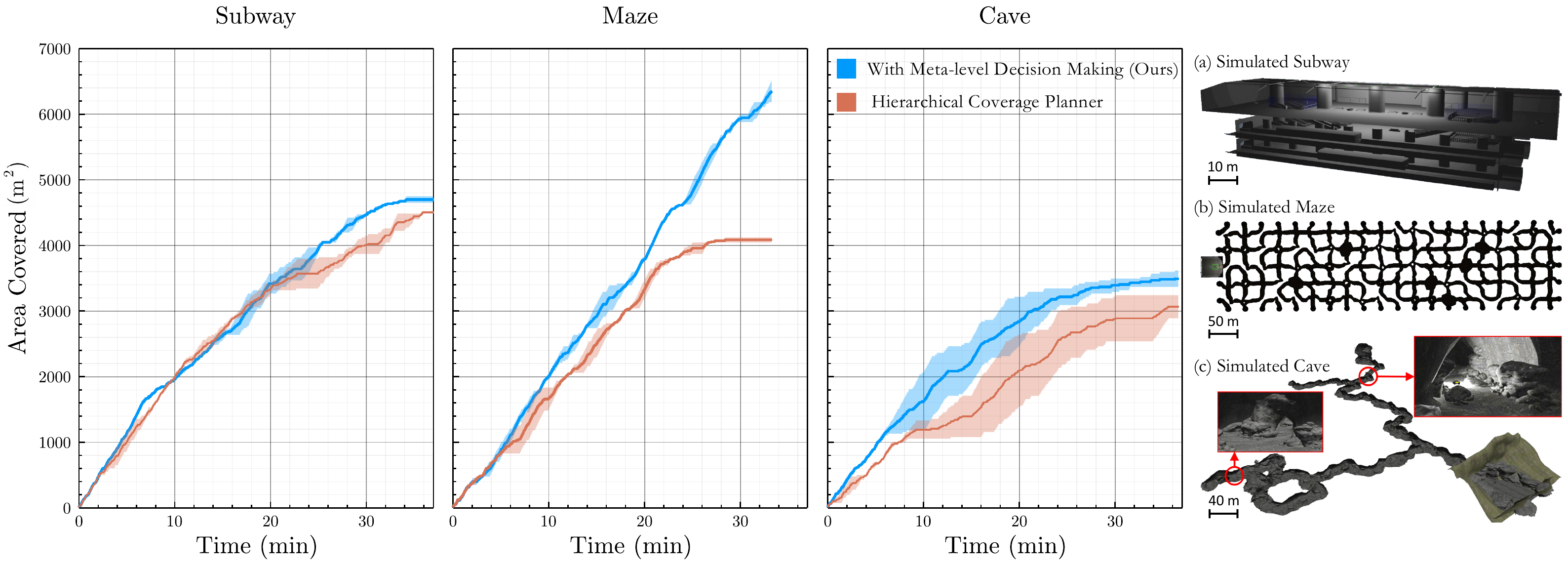}}
\caption{Results showing the exploration performance with and without the meta-level decision making algorithm in the simulated subway, maze, and cave environments. The covered area is the average of two runs and the bounds denote maximum and minimum values between the runs.}
\label{sim_results}
\end{figure}

\begin{figure*}[t]
\centerline{\includegraphics[width=0.81\textwidth]{  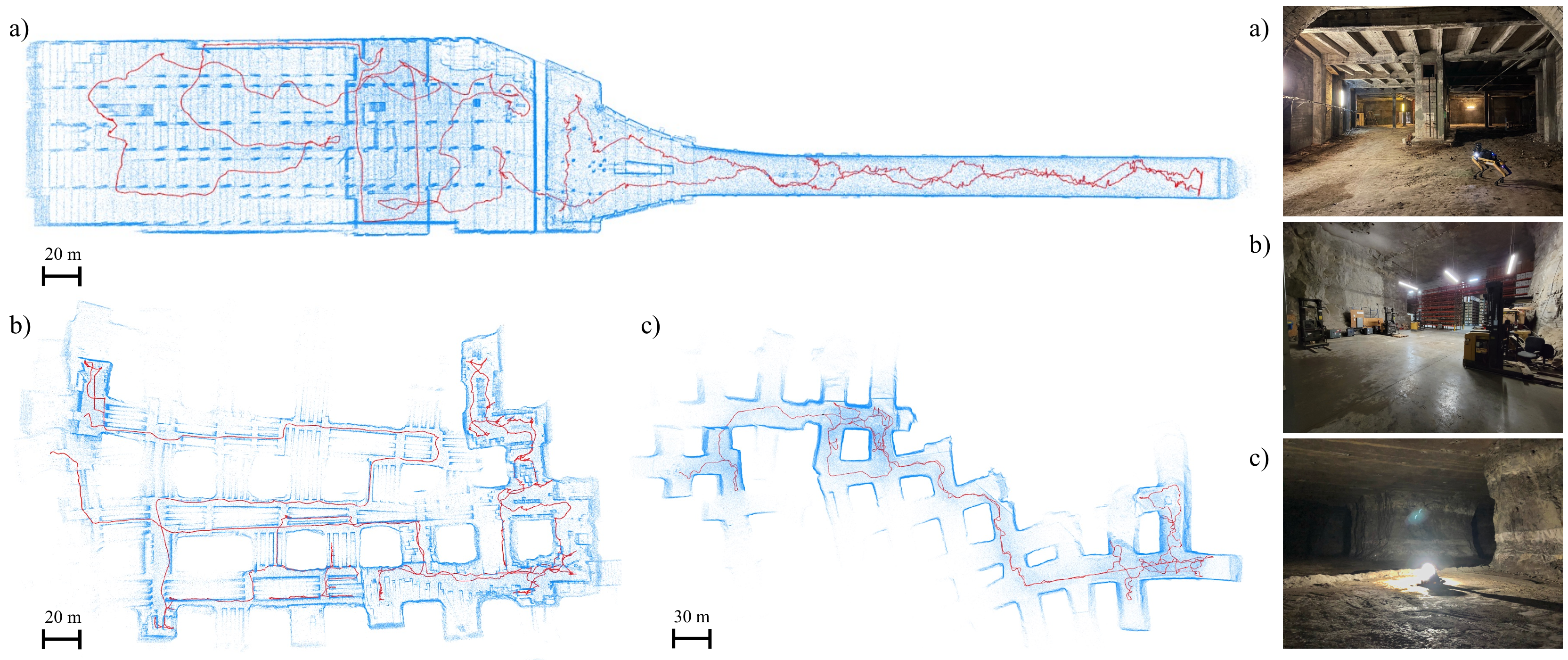}}
\caption{(a) Map of the underground Los Angeles Subway Station, (b) Kentucky Underground Storage Facility, and (c) Kentucky Underground Limestone Mines generated during autonomous exploration runs with meta-level decision making. The robot's path is shown in red with the LiDAR point cloud shown in blue.%
}
\vspace{-5mm}
\label{environments}
\end{figure*}

\textcolor{black}{\ph{Global to Local}
A common implementation for global coverage planning is that once a global frontier is selected, the robot navigates to the frontier and must come within a certain distance of that frontier node before it can switch back to local coverage planning.} However, this implementation is an overly conservative constraint. A more efficient implementation is to allow for the global frontier to be overridden and switch to the local planner on the way to the global frontier. Allowing the robot to follow the risk-informed local coverage planner is often preferred in order to explore the uncovered areas closer to the robot. Doing so prevents the robot from traveling a greater distance to a global frontier and then returning to the previous local area to explore the uncovered area that was there originally. An example of this scenario from a real world exploration mission is shown in Figure \ref{glp_override}.

\textcolor{black}{\ph{Local to Global}
The second situation occurs when switching from the local to global policy is more valuable. Hierarchical coverage planners typically give preference to the local coverage planner when a local path is available. However, there are certain situations where the tradeoff between local and global exploration provides us with a global frontier that is more valuable than the current local policy.} An example of this scenario, shown in Figure \ref{mlp_override}, comes from a real world test conducted at the Los Angeles Subway Station.

\ph{Simulation Setup}
To allow for better analysis between consecutive runs, we tested our coverage planning solution extensively in simulated environments, including simulated subway, maze, and
cave environments (Figure \ref{sim_results}). The subway station consists of large interconnected, polygonal rooms with smooth floors, devoid of obstacles. The maze and cave are both unstructured environments with complex terrain (rocks, steep slopes, and sharp drop-offs) and topology (narrow passages, sharp bends, dead-ends, and open spaces).

The results from multiple simulation runs with and without the meta-level decision making considerations are shown in Figure \ref{sim_results}. Note that we are comparing to a hierarchical coverage planner, like that described by \citet{kim2021plgrim}, which achieves state of the art performance. \textcolor{black}{These results show that in environments with complex terrain (like that of the simulated cave), meta-level decision making allows the robot to find alternative routes when it otherwise might get stuck (similar to the example shown in Figure \ref{mlp_override}).} %

\begin{table}[t]
\caption{} 
\resizebox{\columnwidth}{!}{
\normalsize 
\begin{tabular}{@{}lllll@{}}
    \toprule
    \multicolumn{5}{ c }{\textbf{Simulated Maze}}{\textbf{Simulated Subway Station}} \\
    \midrule
    Method & Rate & 20 min Coverage & Rate & 20 min Coverage\\ 
    & (\si{\meter\squared\per\minute}) & (\si{\meter\squared})  & (\si{\meter\squared\per\minute}) & (\si{\meter\squared}) \\ \midrule
    \textbf{MLDM} & \textbf{191.1} & \textbf{3821} & \textbf{170.5} & \textbf{3410}\\ 
    HCP & 165.8 & 3315 & 167.4 & 3347\\ 
    NBV & 41.4 & 827 & 125.6 & 2511\\
    HFE & 71.8 & 1436 & 156.7 & 3133\\
    \bottomrule
    \end{tabular}}
   
\label{table:maze_results}
\end{table}

\ph{Baseline Algorithms}
We compared our meta-level decision making framework against a local coverage planner baseline (next-best-view method), a global coverage planner baseline (frontier-based method), and a hierarchical coverage planner (HCP) like that of \citet{kim2021plgrim}. The results from these baseline comparisons are summarized in Table \ref{table:maze_results} for the simulated maze and subway environments. 
\begin{enumerate}
    \item Next-Best-View (NBV): NBV samples viewpoints in a neighborhood of the robot, and then plans a deterministic path to each viewpoint \cite{bircher2016receding}. The policy search space is the set of viewpoint paths. NBV selects the policy with the maximum reward based on action cost and information gain from the world representation. %
    
    \item Hierarchical Frontier-based Exploration (HFE): Frontier based exploration methods construct a global representation of the world, where frontiers encode approximate local information gain. The policy search space is the set of frontiers. Exploration combines a one-step look-ahead frontier selection with the creation of new frontiers until all frontiers have been explored. %
\end{enumerate}

\ph{Real World Tests}
Finally, in addition to the simulated tests and the two real world examples presented in Figures \ref{glp_override} and \ref{mlp_override}, we also extensively validated our solution on large-scale hardware tests in a variety of subterranean environments shown in Figure \ref{environments}. The main field testing environments were the Kentucky Underground Limestone Mine and the Los Angeles Subway Station. %
Results from three different real world tests that we ran on both Husky and Spot platforms are shown in Figure \ref{environments}. These three environments are very similar to the simulated subway, cave, and maze domains and demonstrate the capabilities of the onboard autonomy for exploring large unknown environments.

%% file: 6_Conclusion.tex
Our contribution is focused on maximizing the environment coverage during exploration using hierarchical planning structures. \textcolor{black}{Specifically, we introduce a method that is able to adaptively switch between local and global policies in real time by estimating the probability of policy success and the reward for successful policy execution. %
By reasoning about the probability of successful policy execution we are able to leverage risk-aware information about the environment to incorporate greater synergy into the interactions of the local and global policies within the hierarchical planning structure.} These methods have all been tested extensively in simulation and on real world physical robots. They outperform existing solutions by covering up to 1.5 times more area in certain environments.

Future work to expand the overall framework of meta-level decision making includes developing communication-aware exploration strategies. \textcolor{black}{These strategies serve a similar purpose to the local and global policy switching, except the communication-focused override sets a goal that brings the robot back into the communication range of the base station and other robots}. A robot that is out of communication range for too long does not contribute to the overall mission. Another promising area of future work is rare event simulation. The cases in which meta-level decision making proves valuable are often situations where the robot would get stuck if it were not accounting for these tradeoffs. Finding these rare events can be very time consuming when working with simulation and hardware and therefore this work would greatly benefit from the use of adaptive stress testing. By collecting isolated and repeatable test cases, the meta-level decision making solution can be further improved on a diverse set of rare events leading to greater overall efficiency and safety. %

%% file: 7_Acknowledgements.tex
This work is supported by the the Jet Propulsion Laboratory, California Institute of Technology, under a contract with NASA (80NM0018D0004), Defense Advanced Research Projects Agency (DARPA), the BRAILLE Mars Analog project funded by the NASA PSTAR program (NNH16ZDA001N), and the Stanford Intelligent Systems Laboratory (SISL). We also wish to thank Lauren Ho-Tseung, Kelly Harrison, Ransalu Senanayake, the Kentucky Underground staff, and the entire CoSTAR Team.